\documentclass[letterpaper, 10 pt, conference]{ieeeconf}  

\IEEEoverridecommandlockouts                              
\usepackage[utf8]{inputenc}
\usepackage[T1]{fontenc}
\usepackage{amssymb}
\usepackage{amsmath}
\usepackage[ruled]{algorithm2e}
\usepackage{amsfonts}
\usepackage{dsfont}
\usepackage{graphicx}   
\usepackage{url}   
\usepackage[backend=biber, style=ieee]{biblatex}
\usepackage{caption}
\usepackage{subcaption} 
\usepackage{stfloats}

\usepackage{xcolor,colortbl}

\definecolor{Gray}{gray}{0.85}
\definecolor{LightCyan}{rgb}{0.88,1,1}

\newcolumntype{a}{>{\columncolor{Gray}}c}
\newcolumntype{b}{>{\columncolor{white}}c}

\title{\LARGE \bf
GOHOME: Graph-Oriented Heatmap Output for future Motion Estimation
}

\author{Thomas Gilles$^{1, 2}$, Stefano Sabatini$^{1}$, Dzmitry Tsishkou$^{1}$, Bogdan Stanciulescu$^{2}$, Fabien Moutarde$^{2}$

\thanks{$^{1}$IoV team, Paris Research Center, Huawei Technologies France}%
\thanks{$^{2}$MINES ParisTech, PSL University, Center for robotics}%
\thanks{Contact: thomas.gilles@mines-paristech.fr}%
}

\begin{document}

\maketitle
\thispagestyle{empty}
\pagestyle{empty}

\begin{abstract}

    In this paper, we propose GOHOME, a method leveraging graph representations of the High Definition Map and sparse projections to generate a heatmap output representing the future position probability distribution for a given agent in a traffic scene. This heatmap output yields an unconstrained 2D grid representation of agent future possible locations, allowing inherent multimodality and a measure of the uncertainty of the prediction. Our graph-oriented model avoids the high computation burden of representing the surrounding context as squared images and processing it with classical CNNs, but focuses instead only on the most probable lanes where the agent could end up in the immediate future. GOHOME reaches 2$nd$ on Argoverse Motion Forecasting Benchmark on the MissRate$_6$ metric while achieving significant speed-up and memory burden diminution compared to Argoverse 1$^{st}$ place method HOME. We also highlight that heatmap output enables multimodal ensembling and improve 1$^{st}$ place MissRate$_6$ by more than 15$\%$ with our best ensemble on Argoverse. Finally, we evaluate and reach state-of-the-art performance on the other trajectory prediction datasets nuScenes and Interaction, demonstrating the generalizability of our method.

\end{abstract}

\begin{figure*}[b]
\centerline{\includegraphics[width=1.95\columnwidth]{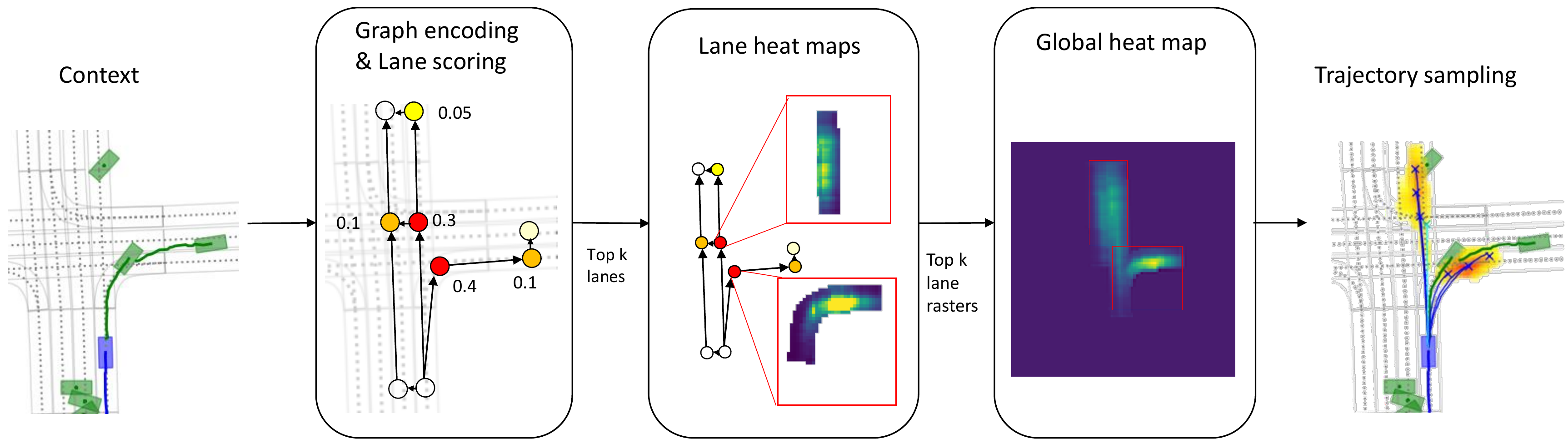}}
\caption{GOHOME pipeline. The lane graph extracted from the HD-Map is processed through a graph encoder. Each lane then generates a local curvilinear raster that is combined into a predicted probability distribution heatmap. }
\label{fig:gohome_pipeline}
\end{figure*}

\section{INTRODUCTION}

Trajectory prediction inherently faces many uncertainties. These uncertainties can be split in two categories: aleatoric and epistemic. Aleatoric uncertainty is the natural randomness of a process: it is the consequence of control noise and will lead to variations in acceleration, curvature, etc ... It translates into a spread of the possible future position and is often tackled by the use of Gaussian predictions in motion estimation \cite{deo2018multi, mercat2020multi}. Epistemic uncertainty outlines some knowledge that can't be known by the observer at prediction time: what is the car destination, will it choose to overtake the car by the left or stay behind ? Recent methods use multimodal outputs based on anchors \cite{chai2020multipath, phan2020covernet}, predefined learning heads \cite{cui2019multimodal} or the available HD-Map \cite{zhao2020tnt, zeng2021lanercnn, narayanan2021divide, kim2021lapred} to cover the span of these possible manoeuvres.
    
    However, existing methods trying to deal with the aforementioned uncertainties in trajectory prediction have limitations. Gaussian predictions are constrained to a 2D predefined shape, that cannot adapt easily to the specific road context, for example at high speed on a curvy road the distribution of the future agent position should resemble the center lane curvature. Regressed sets of coordinates may encounter mode collapse and converge to the same solution, as they are trained with only one ground truth per sample. On the other hand, anchor-based and map-based predictions are restricted to a predefined set of possibilities, depending on preprocessed trajectory clustering or a fixed sampling of the centerlines.
	
	Motion forecasting can also be tackled through the use of a heatmap output representing the final trajectory point location distribution \cite{gilles2021home, mangalam2020goals}. The intermediary waypoints can then be reconstructed  from the history and end point. This brings many advantages for uncertainty modelization and multimodal prediction. First of all it does not restrict the representation of the prediction uncertainty to a parametric form (like a gaussian). Moreover, it conveys a richer information regarding the future probability distribution compared to a predefined number of predicted trajectories, enabling predictions with a much better coverage . This is usually achieved rasterizing an image to represent the context around the car, and processing it through an encoder-decoder CNN. However, the distance a car can travel in a given time horizon can exceed this image boundary, and extending the reach results in quadratic complexity increase. Moreover, convolutional networks commonly used for the task of generating images have to operate on the full square image while the actual road and drivable area take a much sparser space.

    HOME \cite{gilles2021home} introduces the use of a probability heatmap as model output for car trajectory prediction, but uses a fully-convolutional model and is limited to a restricted image size . We iterate on this work and present an optimized motion forecasting framework solely based on graph operations to provide efficiently an heatmap with uncertainty measure exploiting the vectorized form of HD map. We also highlight that heatmap output is suitable for model ensembling without any risk of mode collapse and bring significant improvement to the state-of-the-art using this ensembling.
    Our GOHOME pipeline is illustrated in Fig. \ref{fig:gohome_pipeline}.

\section{RELATED WORK}

\label{sec:citations}

The sequential nature of temporal trajectories makes them a straightforward application of recurrent neural networks \cite{altche2017lstm, mercat2020multi}. However, the need for local map and context information leads them to be often combined with Convolution Neural Networks (CNNs) applied on top-view images \cite{lee2017desire, tang2019multiple, cui2019multimodal, liu2021multimodal}.


\begin{figure*}[b]
\centerline{\includegraphics[width=2.0\columnwidth]{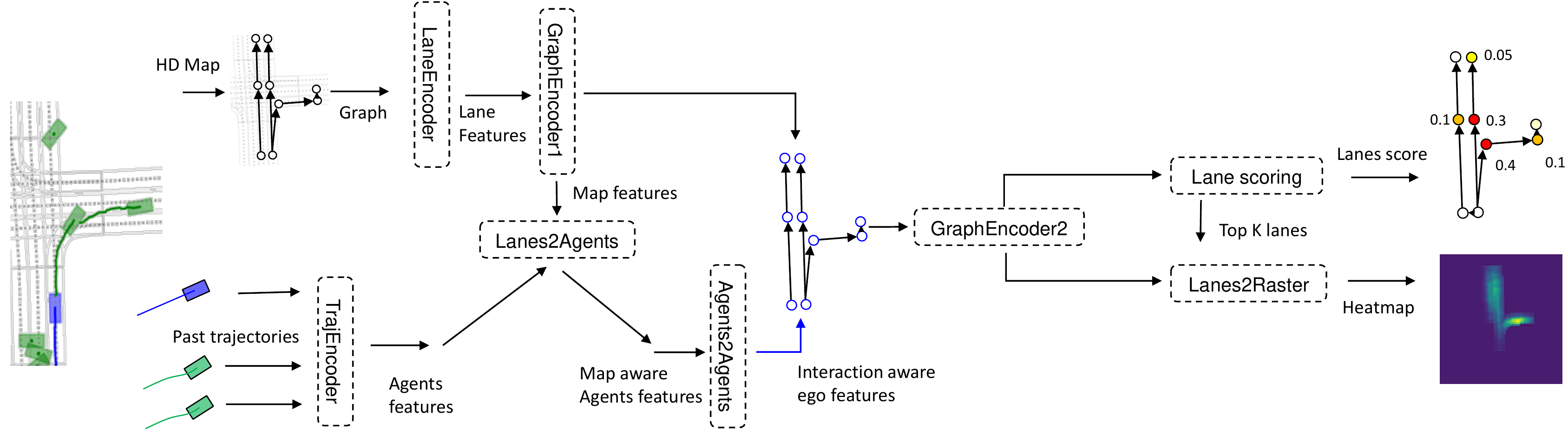}}
\caption{GOHOME model architecture}
\label{fig:gohome_architecture}
\end{figure*}

Lately, Graph Neural Networks (GNNs) have been increasingly applied in order to process compact map encodings with a deepened connectivity understanding. VectorNet \cite{gao2020vectornet} treats indifferently trajectory and map lanes as sets of points (polylines), and encodes them into a global interaction graph. LaneGCN \cite{liang2020learning} uses graph convolutions onto the connected lane graph before fusing this lane information with actor information. 

While most previous works tackle multimodality through learned regression heads, recent work brought different output representations in order to avoid mode collapse and sample inefficiency. CoverNet \cite{phan2020covernet} and MultiPath \cite{chai2020multipath} use anchor trajectory priors to have identified modes without risk of averaging them. PRIME \cite{song2021learning} generates model-based trajectories and then ranks them with a learned model. TNT \cite{zhao2020tnt} samples target candidates along lane priors and scores them using a VectorNet backbone, while LaneRCNN \cite{zeng2021lanercnn} generates a lane graph for each actor and uses the lane nodes as a classification output for future position. GoalNet \cite{zhang2020map} identifies possible long-term goals proposals with the map, and runs a GNN where the features for each possible goal are a path-relative raster.

Generative methods can also be used to obtain multimodal predictions, through either Variational Auto Encoders \cite{lee2017desire, rhinehart2018r2p2, tang2019multiple, mangalam2020not} or Generative Adversarial Networks \cite{alahi2016social, sadeghian2019sophie}, but they require multiple forward passes for each prediction, do not guarantee diversity in the samples obtained and their inherent randomness in not advisable in production systems..

Other methods focus on a heatmap output to represent the future distribution, as it possesses natural multimodality and is therefore not subject to mode collapse. Mangalam \textit{et al.} \cite{mangalam2020goals} models both long-term goals and intermediary waypoints in the two-dimensional space as an image for pedestrian trajectory forecasting, combined with random sampling and Kmeans clustering. HOME \cite{gilles2021home} predicts a future probability heatmap for car trajectories, and devises deterministic sampling algorithms for various metric optimization.
However, most of these heatmap-based methods use a full CNN architecture. To our knowledge, we are the first work to combine a GNN architecture and a heatmap output without the use of any CNNs.

\section{METHOD}
\label{sec:method}

The goal is to output a heatmap that represents the position of an agent at a given time in the future. The trajectory is then regressed conditioned to the final end point.
To achieve this, our GOHOME system focuses on lane-level operations as illustrated in Fig. \ref{fig:gohome_pipeline}.
The local HD-Map is provided as a graph of $L$ lanelets. A lanelet represents a macro section of the road (10 to 20 meters on average), as our goal is to encode connectivity at a macro level (lane segments), and not micro level (every meter). Each lanelet is defined as a sequence of centerline points, and is connected to its predecessor, successor, left and right neighbors if they exist. 
We encode each lanelet into a road graph, where geometric and connectivity information are represented. Our model yields a score for each of these lanelets, that is used to identify most probables lanes. A partial heatmap is then generated for the top ranking lanelets, and projected onto a global heatmap. Afterwards, we sample a set of endpoints from the heatmap and recreate a trajectory for each.

\subsection{Graph neural network for HD-Map input}

The model architecture is illustrated in Fig. \ref{fig:gohome_architecture}. We encode each lanelet through a shared 1D convolution and UGRU \cite{ugru, rozenberg2021asymmetrical} recurrent $LaneEncoder$ into features $F$ of $C$ channels. The lanelet features $F$ are then updated through a $GraphEncoder1$ made of a sequence of four graph convolution operations similar to \cite{liang2020learning} in order to spread connectivity information:
\begin{equation}
    F \leftarrow F W + \sum_r A_r F W_r
\end{equation}
where $F$ is the $(L, C)$ lane feature matrix , $W$ is the learned $(C, C)$ weight for ego features encoding. $A_r$ and $W_r$ are the respective adjacency matrix $(L,L)$ and learned weight $(C,C)$ for the relation $r \in \{predecessor, successor, left, right \}$ derived from the lane graph. $A_r$ is fixed as it comes from the HD Map, while  $W_r$ enables to learn different operations for each possible relation.

Parallely, each agent trajectory, defined as a sequence of position, speed and yaw, is encoded with a shared $TrajEncoder$ also made of a shared 1D convolution and a UGRU layer. Each agent feature is then updated with map information through a cross-attention $Lanes2Agents$ layer on the lanelet features. Interactions between agents are then taken into account through a self-attention $Agents2Agents$ layer between agents.  
Finally the target agent feature is concatenated to all the lanelet features by $Ego2Agents$ and then treated through a final $GraphEncoder2$ layer, also made of 4 graph convolutions to obtain the final graph encoding that will be used to generate the different predictions.

Compared to other methods using graph neural networks, our method uses graph convolutions like LaneGCN \cite{liang2020learning} and LaneRCNN \cite{zeng2021lanercnn}, but applies them to lanelets instead of lane nodes (a lane node is a single point in the sequence of a lanelet). VectorNet \cite{gao2020vectornet} and TNT \cite{zhao2020tnt} also use lanelets, called polylines, but connect them through global attention instead of using graph connectivity. We chose to use a GNN on the lanelets since we wanted an efficient and high-level representation allowing to spread information easily through the graph, while still leveraging connectivity. 

\subsection{Heatmap generation through Lane-level rasters}

For the heatmap output, we wish to have a dense image in cartesian coordinates of dimensions $(H, W)$. To do so without using any convolution on the full image, we create a raster for every lanelet in curvilinear coordinates. We use lane ranking to generate these lane rasters only for the top $k$ lanelets and not all of them.

\begin{figure}[h]
\centerline{\includegraphics[width=1\columnwidth]{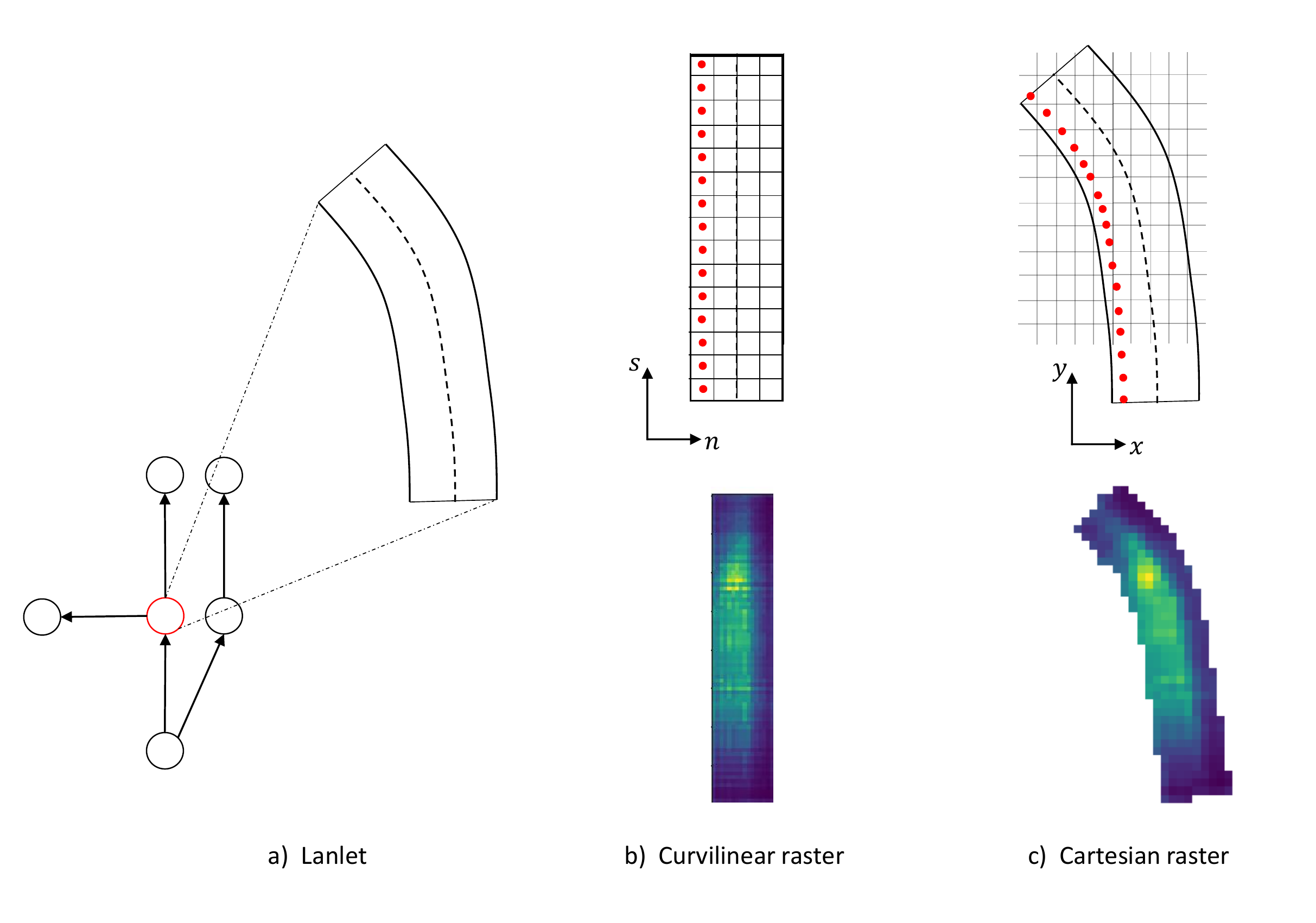}}
\caption{Lane raster grid projection onto cartesian coordinates. a) A single node of the graph is a lanelet and describes a road segment. b) A rectangular raster is generated along the curvilinear coordinates of the lanelet. c) The lanelet coordinates are then used to project the predicted raster back into cartesian coordinates to complete the final heatmap output.}
\label{fig:laneraster_proj}
\end{figure}

\paragraph{Lane raster generation} Each of the small lane rasters of size $(h, w)$ has a longitudinal lenght of 20m and a transversal width of 4m. These lane rasters are created as a discretization of the Frenet-Serret referential along the lane , as illustrated in Fig. \ref{fig:laneraster_proj}.
We decompose the probability distribution along a lanelet in a longitudinal component $(h,1,8)$ and a lateral component $(1,w,8)$ predicted from the lanelet encoding. These components are summed together with broadcast to create a $(h ,w, 8)$ $R_{features}$ volume. This way the complexity to create the volume is $(h+w)\times8$ instead of $h\times w\times8$. The obtained volume is then concatenated with pixelwise cartesian coordinates, heading, lane occupancy and curvature informations before a final linear layer on which is applied a sigmoid to get the $R_{proba}$ output.

The resulting lane-level $R_{proba}$ heatmaps are then projected onto the full cartesian heatmap $\hat{Y}$ using each pixel cartesian coordinates as illustrated in Figure \ref{fig:laneraster_proj}. If multiple lane-level pixels fall into the same cartesian pixel, their values are averaged. The target $Y$ for this final prediction $\hat{Y}$ is a Gaussian centered around the target agent ground truth position. We use the same pixel-wise focal loss as \cite{gilles2021home} inspired from CenterNet \cite{zhou2019objects}:

\begin{equation}
\begin{split}
L = -\frac{1}{P}\sum_p (Y_p-\hat{Y}_p)^2 f(Y_p,\hat{Y}_p) \\
\text{with } f(Y_p,\hat{Y}_p)=
    \begin{cases}
      \log(\hat{Y}_p) & \text{if $Y_p$=1}\\
      (1-Y_p)^4\log(1-\hat{Y}_p) & \text{otherwise}
    \end{cases}  
\end{split}
\label{eq:loss}
\end{equation}

\paragraph{Lane ranking}
The lanelet classification is obtained with a linear layer on the graph encoding, followed by a sigmoid activation. The ground-truth is defined by a 1 for all lanelets where the future car position is inside the lanelet polygon and 0 otherwise. The loss is a binary cross-entropy added to the pixel-wise loss of Eq. \ref{eq:loss} with a $1e^{-2}$ coefficient.
Since only a fraction of the lanelets will actually be useful to represent the future car location, we can compute the lane-level rasters only for a subselection of lanelets, saving more computation. We use the classification score $c_{lane}$ predicted by the network to select only the top $k$ ranking lanelets, and only compute and project the lane raster for these. Since the raster predictions for the other unselected lanelets would have been very close to zero anyway, this does not decrease performance at all, as demonstrated in Sec. \ref{sec:abl_ranking}.

\paragraph{Cartesian image connection} Some lane rasters may be projected onto the same pixels and overlap, which is very difficult for the model to know of beforehand. 
To help the model know of overlaps and propagate location information through the lane rasters, we do a first projection of the lane rasters onto the cartesian coordinates before the final probability estimation. The $(h ,w, 8)$ lane raster features $R_{features}$ are projected onto a $(H ,W, 8)$ cartesian image $I_{features}$ through the same operation previously described for the probability heatmaps, with the overlaps averaged. The occupancy (number of raster pixels aggregated in each cartesian pixel) information is also concatenated to the volume  $I_{features}$. A linear layer is then applied on the last dimension of $I_{features}$, which is then reprojected onto curvilinear coordinates and concatenated to the initial lane rasters $R_{features}$ before final probability estimation and projection. This way, the features of overlapping lanes are shared between them so that they can propagate information and homogenise probability.

\begin{table*}[b]
\begin{minipage}{.5\linewidth}
\caption{Argoverse Leaderboard \cite{leaderboard}}
    \begin{center}
    \begin{tabular}{l|c c|c c a}
      \hline
       & \multicolumn{2}{c|}{K=1}  & \multicolumn{3}{c}{K=6}  \\
      & minFDE& MR&minADE & minFDE & MR \\
      \hline
      LaneGCN \cite{liang2020learning} & 3.78 & 59.1 &  0.87 & 1.36 & 16.3\\
      TPCN \cite{ye2021tpcn}  & 3.64 & 58.6 &  \textbf{0.85} & 1.35  & 15.9\\
      Jean \cite{mercat2020multi}  & 4.24 & 68.6 &  1.00 & 1.42 & 13.1\\
      SceneTrans \cite{ngiam2021scene} & 4.06 & 59.2 & \textbf{0.80} & \textbf{1.23} & 12.6 \\
      LaneRCNN \cite{zeng2021lanercnn} & 3.69 & 56.9 &  0.90 & 1.45  & 12.3\\
      PRIME  \cite{song2021learning}& 3.82 & 58.7 &  1.22 & 1.56   & 11.5\\
      DenseTNT \cite{gu2021densetnt} & 3.70 & 59.9 &  0.94 & 1.49 & 10.5\\
      HOME \cite{gilles2021home}  & 3.65 & 57.1 &  0.93 & 1.44   & 9.8\\
      \hline
      GOHOME & \textcolor{black}{3.65} & \textcolor{black}{57.2} &  \textcolor{black}{0.94} & \textcolor{black}{1.45}  & \textcolor{black}{10.5}\\
      \hline
      HO+HO & 3.57 & 56 &  0.92 & 1.41   & 9.4\\
      HO+GO  & \textbf{3.53} & \textbf{55.8} &  0.92 & 1.40 & 9.1\\
      Best ensemble & 3.68 & 57.2 &  0.89 & 1.29 & \textbf{8.5}\\
      
      \hline
    \end{tabular}
    \end{center}
    \label{tab:argo_test}
\end{minipage}
\begin{minipage}{.5\linewidth}
\caption{NuScenes Leaderboard \cite{nuscenesleaderboard}}
    \begin{center}
    \begin{tabular}{l|c c|c c| c}
      
      \hline
      & \multicolumn{2}{c|}{K=5}  & \multicolumn{2}{c}{K=10} & k=1  \\
      & minADE & MR & minADE  & MR & minFDE \\
      \hline
       CoverNet \cite{phan2020covernet}             & 1.96 & 67 &  1.48  & \_ & \_\\
      Trajectron++ \cite{salzmann2020trajectron++} & 1.88 & 70 &  1.51  & 57 & 9.52\\
      ALAN \cite{narayanan2021divide}              & 1.87 & 60 &  1.22  & 49 & 9.98\\
      SG-Net \cite{wang2021stepwise}               & 1.86 & 67 &  1.40  & 52 & 9.25\\
      WIMP \cite{khandelwal2020if}                 & 1.84 & 55 &  \textbf{1.11}  & 43 & 8.49\\
      MHA-JAM \cite{messaoud2020multi}             & 1.81 & 59 &  1.24  & 46 & 8.57\\
      CXX \cite{luo2020probabilistic}              & 1.63 & 69 &  1.29  & 60 & 8.86\\
      LaPred \cite{kim2021lapred}                  & 1.53 & \_ &  1.12  & \_ & 8.12\\
      P2T \cite{deo2020trajectory}                 & 1.45 & 64 &  1.16  & 46 & 10.50\\
      \hline
      GOHOME (r=2.6m)                                       & \textbf{1.42} & 57 &  1.15 & 47 &  \textbf{6.99}\\
      GOHOME (r=1.8m)                              & 1.59 & \textbf{46} &  1.15  & \textbf{34} & 7.01\\
      \hline

      \hline
    \end{tabular}
    \end{center}
    \label{tab:nuscenes_test}
\end{minipage}
\end{table*}

\begin{table*}[t]
\begin{minipage}{.3\linewidth}
\captionof{table}{Interaction Validation}
\begin{center}
\begin{tabular}{l|c c}
  \hline
      & minFDE$_1$ & minFDE$_6$ \\
  \hline
  TNT \cite{zhao2020tnt} &  \_ & 0.67\\
  HEAT-I-R \cite{mo2021heterogeneous} & 0.66 & \_ \\
  ReCoG \cite{mo2020recog}  &   0.65 & \_ \\
  ITRA \cite{scibior2021imagining} & \_ & 0.49 \\
  GOHOME   &  \textbf{0.61} & \textbf{0.45} \\
  \hline
\end{tabular}
\end{center}
\label{tab:interaction}
\end{minipage}
\begin{minipage}{.4\linewidth}
\captionof{table}{Performance/Complexity comparison}
\begin{center}
\begin{tabular}{l|c c|c| c}
  \hline
   Model   & \multicolumn{2}{c|}{K=6} & $\#$Param & FLOPs \\
  &  minFDE& MR &  \\
  \hline
  HOME &  1.28 & 6.8 & 5.1M  &  4.8G\\
  GNN-HOME  &   1.28 & 7.2 & 0.43M   & 0.81G\\
  GOHOME   &  1.26 & 7.1 & 0.40M & 0.09G \\
  \hline
\end{tabular}
\end{center}
\label{tab:gnn_speedup}
\end{minipage}
\begin{minipage}{.3\linewidth}
\captionof{table}{Lane ranking speed-up}
\begin{center}
\begin{tabular}{l|c c|c}
  \hline
   $\#$ lanes    & \multicolumn{2}{c|}{K=6} & FPS \\
   & minFDE& MR &  \\
  \hline
  All  &  1.28 & 7.5 & 17\\
  20   &   1.26 & 7.1 & 34\\
  10   &  1.26 & 7.3 & 45\\
  \hline
\end{tabular}
\end{center}
\label{tab:rank_speedup}
\end{minipage}

\end{table*}


\subsection{Sparse sampling for Miss Rate Optimization and Full trajectory generation}

To derive final trajectory points from the heatmap, we use the same MissRate optimization sampling algorithm as HOME \cite{gilles2021home}: we iteratively select the grid point that maximizes surrounding probability in a local neighborhood of radius $r$, then set this local neighborhood probabilities to zero so the next iteration doesn't select the same location. The radius $r$ is a simple hyper-parameter that can be tuned according to the spread of uncertainty present in the dataset, as we will demonstrate in Sec. \ref{sec:sota}.

Full trajectories are then inferred from the sampled end points with a simple fully connected model of 2 hidden layers with 64 features each, taking the history and end point as inputs and trained with the ground truth end points.

\subsection{Model ensembling}

Because of the multimodal nature of the predictions, model ensembling is usually not possible in trajectory prediction. It is indeed not possible to determine which modality should be averaged with which, and even shortest distance matching doesn't guarantee that the two predictions highlight the same decision and would make sense averaged together. On the other hand, probability heatmap are a great way of representing information coming from different sources or models in a common system of reference and can be averaged together without any assumption nor risk of mode collapsing. 

\section{Experimental Results}
\label{sec:result}

\subsection{Experimental settings}

\paragraph{Datasets}
The Argoverse Motion Forecasting Dataset \cite{chang2019argoverse} is made of 205942 training samples, 39472 validation samples and 78143 test samples, with 2 seconds history, and 3 seconds future sampled at 10Hz. The NuScenes Prediction dataset \cite{caesar2020nuscenes} is made of 32186 training samples and 9041 validation samples, with 2 seconds history and 6 seconds future sampled at 2Hz. The Interaction dataset is made of 447626 training samples and 130403, with 1 seconds history and 3 seconds future trajectory sampled at 10Hz. All datasets provide the local HD-Map as a lanelet graph.

\paragraph{Metrics} The metrics commonly used by both datasets are the MR$_k$ and minFDE$_k$, which are respectively the Miss Rate and the minimum Final Displacement Error for the top $k$ predictions, as well as the minimum Average Displacement Error minADE$_k$. Following their respective leaderboards, we report these metrics for k=1,5 for Argoverse, k=1,5,10 for NuScenes and k=1,6 for Interaction.

\paragraph{Implementation details} All models are trained with batchsize 32 and Adam optimizer. Trainings last 16 epochs for validation evaluation and 32 epochs for test evaluation. The initial learning rate is 1e$^{-3}$ and is divided by 2 at epochs 3, 6, 9 and 13. All layers have 64 channels, graph convolution and attention layers are followed by Layer Normalization \cite{ba2016layer}. ReLU activation is used after every layer.
We upsample the HD-Maps lanelets to obtain an average lenght of 10m per lanelet. 

The architecture is the same for all datasets, with the exception of the sampling radius $r$ that we can tune according to the uncertainty spread of the dataset and the metrics we want to optimize.  The default sampling radius we use for Argoverse is 1.8m, slightly less than the 2m threshold defining the MissRate. Since the nuScenes dataset has a longer prediction horizon, it is more uncertain  and therefore a larger radius of 2.6m is required to optimiwe the minADE$_5$, however MissRate remains better for a lower radius of 1.8m. Finally, since Interaction perception data comes from drones, it is much less noisy and therefore generates much more focused probability predictions which we sample with a radius of 1.4m.

\subsection{Comparison with State-of-the-art}
\label{sec:sota}

We report our results on the online Argoverse test set in Tab. \ref{tab:argo_test}, on the online NuScenes leaderboard in Tab. \ref{tab:nuscenes_test}, and on the Interaction validation set in Tab. \ref{tab:interaction}. We compare it to the published methods on each of these benchmarks.
On Argoverse, our GOHOME method reaches 2$^{nd}$ place in MR$_6$, with the use of a lighter and faster model than 1$^{st}$ HOME as will be showed in Sec. \ref{sec:ablation_studies}. 
On NuScenes and Interaction, GOHOME ranks first in multiple metrics as well.


\begin{figure*}[b]
    \center
    \begin{minipage}[b]{0.45\textwidth}
        \includegraphics[width=\textwidth]{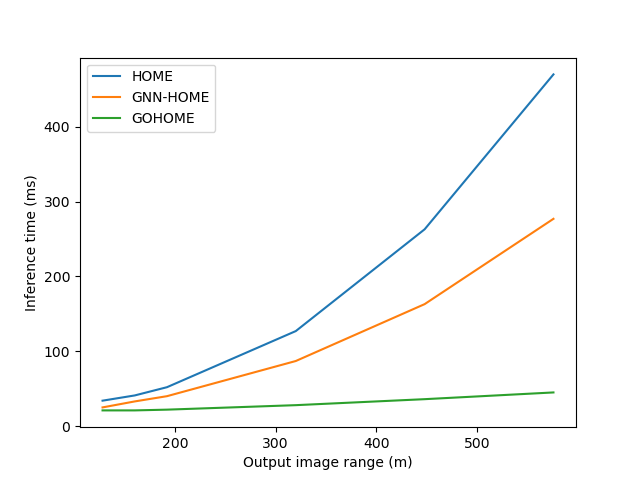}
        \caption{Inference time with regard to output range}
        \label{fig:out_scale}
    \end{minipage}
    \begin{minipage}[b]{0.45\textwidth}
        \includegraphics[width=\textwidth]{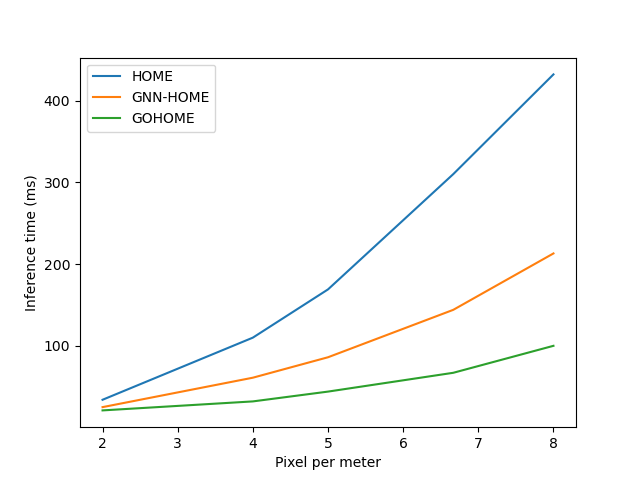}
        \caption{Inference time with regard to pixels per meters}
        \label{fig:pixel_scale}
    \end{minipage}
    \label{fig:resdi}
\end{figure*}

\paragraph{Ensembling for increased performance and highlight of model differences} We also report the results of our ensembled models on the test set. We first highlight that the ensemble of two similar HOME models (HO+HO) brings significant improvement compared to HOME alone. As a general rule, the more different and complementary two models are, the greater the performance increase will be. We notice that the combination of HOME and GOHOME models (HO+GO) brings a greater improvement than HO+HO, despite each HOME model being better in single performance than the GOHOME model. 
Our best ensembling, a weighted combination of 9 HOME and GOHOME models, allows us to improve on the existing state-of-the-art by a significant margin, with a more than 15$\%$ MR$_6$ decrease.


\subsection{Ablation studies}
\label{sec:ablation_studies}

We highlight the gains made by replacing convolution operations with graph operations. To measure inference time, we use a batchsize of 16, which can be considered as an average number of agents to be predicted at a given time. We report only the model forward pass, omitting preprocessing and postprocessing, but notice that image preprocessing is sensibly slower, particularly because of the rasterization of the different semantic layers. All times are measured with a Nvidia 2080 TI. While we report inference time, we also notice that training times record an even greater difference. We mostly consider three different architectures: HOME, GNN-HOME which is a modified HOME model with a GNN encoder but the usual CNN decoder, and our new method GOHOME. All numbers are reported on the Argoverse validation set.

\subsubsection{Graph operation speed-up}

We evaluate the speed-up gained by using graph encoding and lane rasters instead of full convolutions in Tab. \ref{tab:gnn_speedup}. The CNN encoding and decoding corresponds to the full HOME model, and the GNN with Lane Rasters (LR) to the GOHOME model. We also estimate the impact from the encoding separately by testing a model with GNN encoding and CNN decoding. We report FLOPs, number of parameters and Frame per Seconds. We measure an average number of 140 lanelets and 10 agents per sample to compute FLOPs.


\subsubsection{Trade-off from lane ranking}
\label{sec:abl_ranking}

We show in Tab. \ref{tab:rank_speedup} the effects of only selecting the top $k$ lanes to extract rasters. We fix an input range of 128 and output range of 192 with 0.5m x 0.5m pixel resolution. We observe that this ranked selection doesn't decrease performance, as limiting the number of projected lanes seems to actually improve the metrics, and brings effective speed-up.



\subsubsection{Image size and resolution scaling}

While a 192m image range, which amount to a 88m reach in each direction, may be sufficient in most urban driving predictions with a time horizon of 3s, other datasets can require predictions up to 6 or 8 seconds \cite{caesar2020nuscenes,ettinger2021large}. There is therefore a need to increase this output range, which can be done without necessarily increasing the input size, as far distances are reached with long straight trajectories that can be easily extrapolated on highways.


\begin{figure*}[b]
\centerline{\includegraphics[width=2.0\columnwidth]{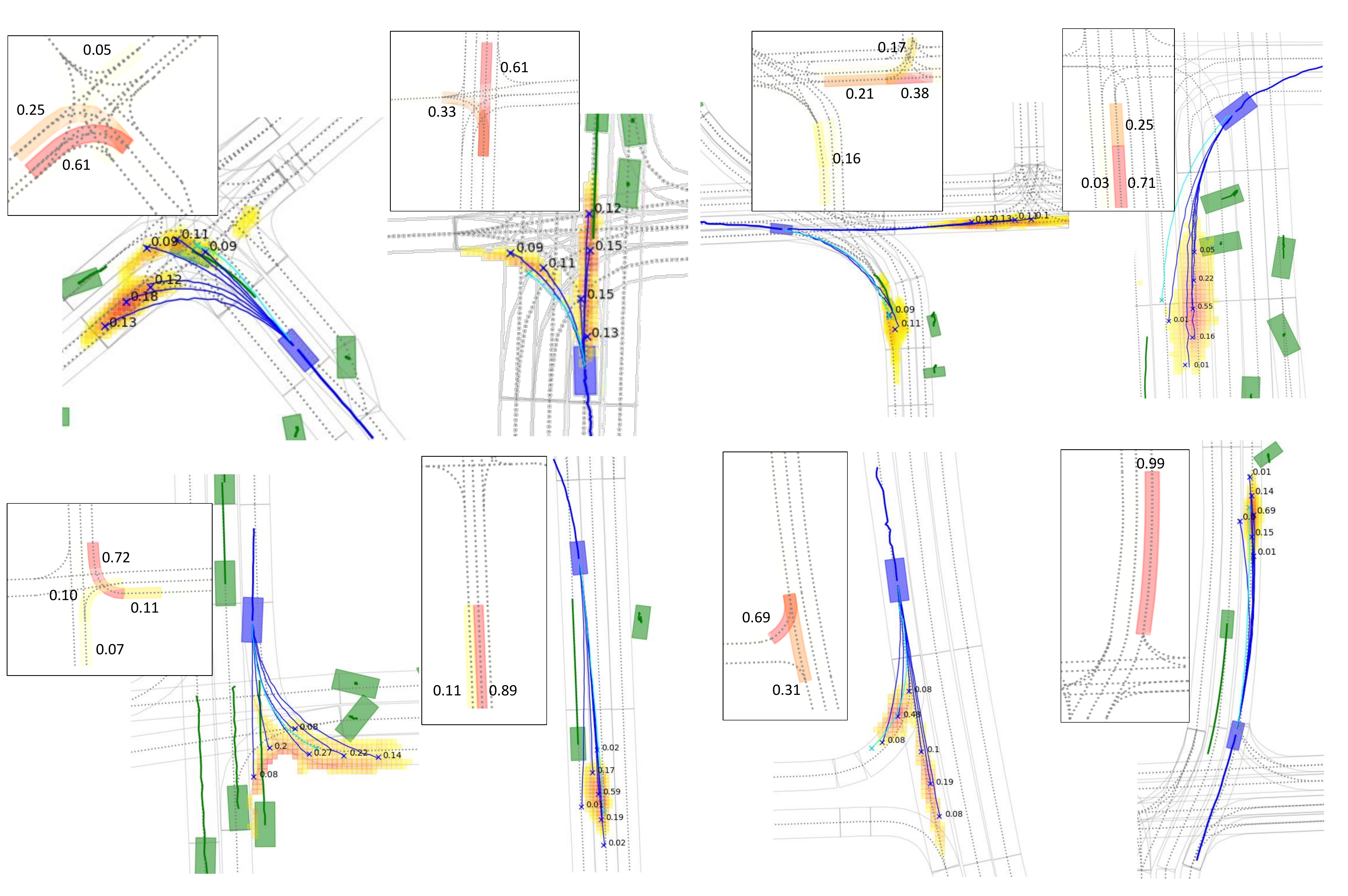}}
\caption{Qualitative examples of GOHOME output. Graph lane classification is shown in framed inserts}
\label{fig:quali}
\end{figure*}

We compare the scaling of our graph-based GOHOME model to the one of an image-based HOME model with regard to output range and resolution. Fig. \ref{fig:out_scale} highlights the output range scaling, where we use a fixed input range of 128 meters, and a resolution of 0.5m per pixel. Whereas the CNN decoders of HOME and GNN-HOME lead to a quadratic scaling, the lane rasters combined with the top 20 lane ranking enable a scaling that is even less than linear.

We show in Fig. \ref{fig:pixel_scale} the inference time with regard to the number of pixels per meter, which is the inverse of the resolution. While efficient optimization of convolution and constant costs lead to close inference times for the initial 2 pixels per meter, the more efficient scaling shows clearly for GNN inputs and especially Lane Rasters outputs. As the resolution of the lane rasters is also scaled with the total resolution, the quadratic complexity is still applied, but with a much lesser coefficient that allows for realistic training and inference times for finer resolutions.

\subsection{Qualitative results}

We show in Fig. \ref{fig:quali} some qualitative results of our GOHOME model. The lane prediction displayed on top can be assimilated to the representation of epistemic uncertainty, as the choice of where the driver will decide to go, whereas the spread of the final heatmap modes models aleatoric uncertainty in the trajectory controls. 
We observe that the model assign different modes for each lane possibility, and that each of these modes is well aligned with the corresponding lane with a spread along the curvilinear direction. 


\section{Conclusion}
\label{sec:conclusion}

In this paper, we propose GOHOME, a trajectory prediction framework generating a global heatmap probability distribution without the use of any image based convolution. Through the use of graph operations, ranking and projections, our model reaches state-of-the-art performance on three datasets with great scaling with regards to the predicted range and resolution. 










\clearpage
\printbibliography

\end{document}